%% file: root.tex
\documentclass[letterpaper, 10 pt, conference]{ieeeconf}  

\IEEEoverridecommandlockouts                              

\usepackage{graphics} 
\usepackage{amsmath} 
\usepackage{import}
\usepackage[detect-all=true,separate-uncertainty=true, multi-part-units=single,range-phrase={-},product-units = single]{siunitx}
\usepackage{menukeys}
\usetikzlibrary{arrows,positioning,shapes.geometric}
\usetikzlibrary{intersections}

\usepackage{amsfonts}
\usepackage[caption=false,font=small]{subfig}
\usepackage{booktabs}

\usepackage[normalem]{ulem}

\title{\LARGE \bf
Collaborative Robotic Biopsy with Trajectory Guidance and \\ Needle Tip Force Feedback
}

\author{Robin~Mieling$^{1}$*,
Maximilian~Neidhardt$^{1}$*, 
Sarah~Latus$^{1}$,
Carolin~Stapper$^{1}$, 
Stefan~Gerlach$^{1}$,\\
Inga~Kniep$^{2}$,
Axel~Heinemann$^{2}$, 
Benjamin~Ondruschka$^{2}$ 
and Alexander~Schlaefer$^{1}$
\thanks{$^{1}$Institute of Medical Technology and Intelligent Systems, Hamburg University of Technology, 21073 Hamburg, Germany 
{\tt\small robin.mieling@tuhh.de}}%
\thanks{$^{2}$Institute of Legal Medicine, University Medical Center Hamburg-Eppendorf, 22529 Hamburg,  Germany}%
\thanks{This research was partially funded by the \mbox{NATON} project (grant agreement no 01KX2121), by the \mbox{Calls for Transfer} initiative (BWFGB Hamburg, C4T535) and the TUHH $i^3$ initiative}%
\thanks{Ethical approval: The Ethics Committee of the Hamburg Chamber of Physicians approved the study (No.: 2020-10353-BO-ff)}%
\thanks{* Both authors contributed equally.}%
}

\newif\ifcopyright
	\copyrighttrue

\usepackage[mathscr]{euscript}

\definecolor{Gray}{gray}{0.85}

\newcolumntype{M}[1]{>{\centering\arraybackslash}m{#1}}
\newcolumntype{N}{@{}m{0pt}@{}}
\makeatletter
\newcommand{\removelatexerror}{\let\@latex@error\@gobble}
\makeatother

\begin{document}
\ifcopyright
{\LARGE IEEE Copyright Notice}
\newline
\fboxrule=0.4pt \fboxsep=3pt

\fbox{\begin{minipage}{1.8\linewidth}  

		Personal use of this material is permitted.  Permission from IEEE must be obtained for all other uses, in any current or future media, including reprinting/republishing this material for advertising or promotional purposes, creating new collective works, for resale or redistribution to servers or lists, or reuse of any copyrighted component of this work in other works. \\
		
		Published in: Proceedings of the 2023 IEEE International Conference on Robotics and Automation (ICRA), May 29 -- June 2, 2023, London, England. \\
  
        The final version of record is available at https://doi.org/10.1109/ICRA48891.2023.10161377. 
		
\end{minipage}}
\else
\fi
\graphicspath{{./images/}}

\maketitle
\thispagestyle{empty}
\pagestyle{empty}

\begin{abstract}

The diagnostic value of biopsies is highly dependent on the placement of needles. Robotic trajectory guidance has been shown to improve needle positioning, but feedback for real-time navigation is limited. Haptic display of needle tip forces can provide rich feedback for needle navigation by enabling localization of tissue structures along the insertion path. We present a collaborative robotic biopsy system that combines trajectory guidance with kinesthetic feedback to assist the physician in needle placement. The robot aligns the needle while the insertion is performed in collaboration with a medical expert who controls the needle position on site. We present a needle design that senses forces at the needle tip based on optical coherence tomography and machine learning for real-time data processing. Our robotic setup allows operators to sense deep tissue interfaces independent of frictional forces to improve needle placement relative to a desired target structure. We first evaluate needle tip force sensing in ex-vivo tissue in a phantom study. We characterize the tip forces during insertions with constant velocity and demonstrate the ability to detect tissue interfaces in a collaborative user study. Participants are able to detect \SI{91}{\ percent} of ex-vivo tissue interfaces based on needle tip force feedback alone. Finally, we demonstrate that even smaller, deep target structures can be accurately sampled by performing post-mortem in situ biopsies of the pancreas.
\end{abstract}

\section{INTRODUCTION}

\begin{figure}[tb]
  \centering
  \includegraphics[width=0.99\columnwidth]{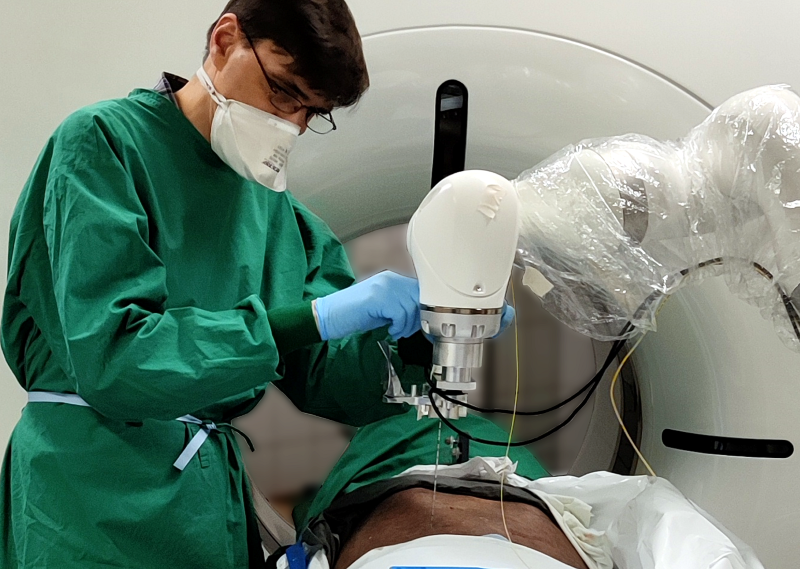}
  \caption{\textbf{Collaborative Robotic Biopsy}: The robot guides the needle trajectory while the physician inserts the needle. Needle tip forces are sensed by the physician through haptic feedback allowing him to feel tissue interfaces during the needle insertion.}
  \label{fig:intro_fig}
\end{figure}

Needles are a valuable tool for reaching soft tissue lesions to extract tissue biopsies for diagnosis or to perform therapy, e.g. radiofrequency ablation or brachytherapy. In clinical practice, needles are placed manually under image guidance, with accuracy depending heavily on the operator~\cite{AtherAdnan.2021,McLeod.2021,Son.2014,Fehrenbach.2021}. Still, tissue sampling relies on precise positioning of the biopsy needle inside the target tissue~\cite{Pritzker.2019}. 
Robotic systems have proven to be beneficial for this task with respect to accuracy, standardization and the number of insertions required~\cite{Siepel.2021}. Robots for needle insertions have been proposed with CT-guidance~\cite{YiyunWangandHongbingLi.,Franckenberg2021SemiautomatedCSFsampling, neidhardt2022robotic} or MRI-guidance \cite{yakar2011feasibility, Vilanova.2020, Futterer.2012}. 

Fully automated insertions have been considered for post-mortem biopsy~\cite{Franckenberg2021SemiautomatedCSFsampling, neidhardt2022robotic} but are still not practical in a clinical environment due to safety concerns. Alternatively, robots align the needle trajectory in the clinic, while needle insertion is performed manually by the physician~\cite{Guiu.2021, Kettenbach.2014, Levy.2021, Martinez.2014}.
Thereby, the physician relies on anatomical knowledge and his or her sense of touch when forwarding the needle into tissue. Similarly, teleoperative systems~\cite{Abdi.2020,aggravi2021haptic,Tai.2021,elayaperumal2014detection, Mendoza.2019, Wartenberg.2018, baksic2021shared} provide haptic feedback to the surgeon when inserting a needle. Forces at the needle tip can be estimated and displayed to the physician~\cite{aggravi2021haptic,elayaperumal2014detection,Han.2018,de2012coaxial} to further enhance the navigation during percutaneous insertions.
Tip forces can be provided by subtracting modeled friction forces from externally measured forces~\cite{aggravi2021haptic,de2012coaxial}. Similarly, miniaturized force sensors can be embedded in needles composed of fiber Bragg gratings~\cite{elayaperumal2014detection, Han.2018}, Fabry Pérot imaging sensors~\cite{Uzun.2020,Beekmans.2016,Su.2011}, or imaging optics~\cite{gessert2019spatio,Ourak.2019}. Needle tip forces have been shown to be superior to forces measured at the shaft for membrane detection~\cite{elayaperumal2014detection, de2012coaxial} but research is limited to phantom studies with synthetic materials and non-commercial robots with small operational spaces and limited degrees of freedom.

In this study, we propose a collaborative robotic system with haptic feedback from the needle tip for minimally invasive tissue biopsies. The physician plans the needle insertion using CT imaging while a lightweight robot aligns the needle according to the planned trajectory. We estimate the ideal trajectory based on the CT-Hounsfield units~\cite{gerlach2021needle,neidhardt2021collaborative}.

Subsequently, needle insertion is performed collaboratively under haptic feedback with both robot and physician on site as shown in Fig.~\ref{fig:intro_fig}. Needle path planning and robot motion are executed in a custom software framework~\cite{neidhardt2022robotic}. We present a \textit{smart} needle with an embedded optical fiber for sensing forces acting on the needle tip. We estimate forces based on optical coherence tomography (OCT) imaging and perform real-time data processing with deep learning~\cite{gessert2019spatio,latus2021rupture,mieling2022proximity}.

The advantages of our system are: (1) safety; the physician can feel critical events and is in control of the robot motion at all times, (2) compensation for displacements and deformations of soft tissue structures during needle insertion~\cite{neidhardt2022robotic, Halstuch.2018, Yang.2018, Jiang.2014, Muthigi.2017} and (3) flexible puncture of different soft tissue targets with a 7 degree of freedom (DOF) robot. 

We evaluate the robotic system in three stages. Firstly, we perform fully robotic insertions with a constant velocity to demonstrate how unpredictable friction forces are in heterogeneous phantoms, and we show that the forces at the needle tip can help identify the location of tissue interfaces. Secondly, we show that operators can determine the topology of the tissue phantoms while purely relying on needle tip force feedback. Lastly, we extract in situ pancreas tissue from corpses which is a challenging target to reach in the clinic requiring CT~\cite{fehrenbach2021ct} or endoscopic ultrasound~\cite{Syed.2019} due to the long insertion path and anatomical localization.  

\begin{figure}[tb]
  \centering
  \includegraphics[width=0.99\columnwidth]{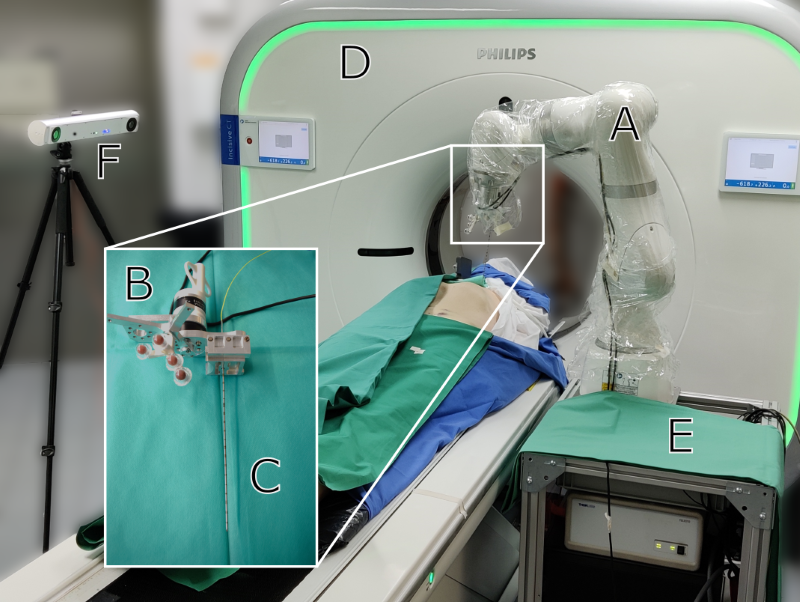}
  \caption{\textbf{Collaborative Robotic System}: The experimental setup consists of the robot (A), the needle mount (B) with integrated force sensors and tracking markers, the \textit{smart} needle (C), the CT system (D), the robot cart (E) with the built-in OCT system and the optical tracking system (F).}
  \label{fig:system_fig}
\end{figure}

\begin{figure}[t]
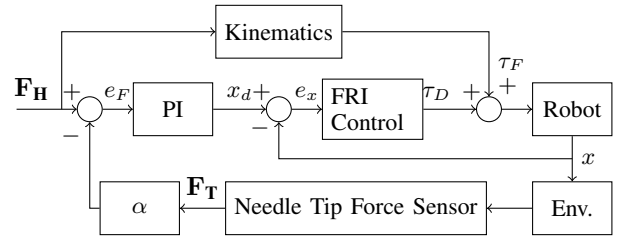

    \centering
    \include{figs/control_scheme}
    \vspace{-0.8cm}
    \caption{\textbf{Control Loop:} Admittance control loop employed to display the forces between needle tip and tissue to the operator. In the outer control loop, the desired robot pose $x_d$ is controlled with a PI controller based on the operator's handle force $\mathbf{F_{\textbf{H}}}$ and the amplified tip force $\mathbf{F_{\textbf{T}}}$. The inner control loop is governed by the position controller of the fast research interface~\cite{schreiber2010fast}. $e_F$ denotes the outer loop force error, $x$ the actual robot pose, $e_x$ the inner loop pose error, $\tau_D$ the controller torque and $\tau_F$ the externally applied torque.}
    \label{fig:control}  
\end{figure}

\section{Methods}

\subsection{Collaborative Robotic System}

\begin{figure*}[tb]
    \centering
	\includegraphics[width=0.97\textwidth]{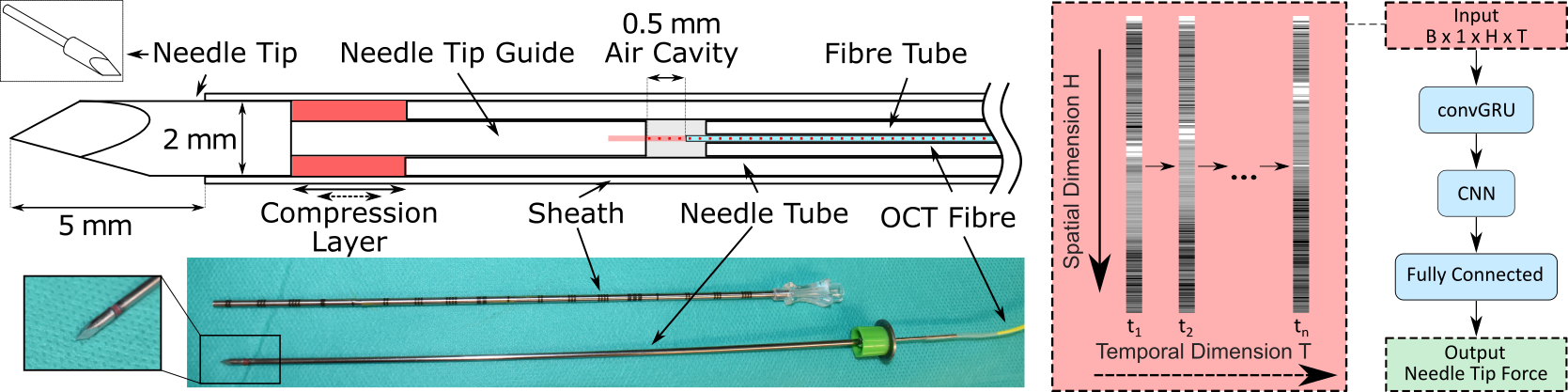}
	\caption{\textbf{\textit{Smart} Needle:} Optical needle probe (left) for the estimation of needle tip forces during the insertion. Forces applied at the needle tip cause the compression layer to deform. The resulting change in air cavity length is resolved in the OCT signal. A symmetric needle tip reduces lateral bending of the needle during insertion. We consider a cGRU-CNN model (right) for real-time processing of the OCT data stream ($\text{H}\times\text{T} = 512\times50$) and compare the performance to a 2D ResNet baseline. The output of the cGRU layer is processed in the regression head containing a ResNet based 1D CNN and fully connected layers.}
	\label{fig:needle}
\end{figure*}

The robotic system is depicted in Fig.~\ref{fig:system_fig} and combines trajectory guidance with haptic needle tip force feedback. The collaborative procedure is performed with a 7-DOF light-weight medical robot (LBR Med 14, KUKA AG, Augsburg, Germany) designed for human-robot interaction. CT images are acquired with the Philips Incisive system and the biopsy target is annotated by medical experts. A custom planning system is used to estimate an entry point prior to the insertion, considering insertion depth, insertion angle and collision avoidance for bone structures. Registration between CT and robot is performed with an optical tracking camera (fusionTrack 500, Atracsys LLC, Puidoux, Switzerland) as described in~\cite{neidhardt2022robotic} and the robot aligns the needle along the chosen trajectory. 
Forces acting on the tip of our \textit{smart} needle are estimated and fed back to the physician as a resistive force. A 3D printed handle allows the surgeon to easily move the \textit{smart} needle along its axis while the robot guides the motion.
Two force-torque sensors (M3703, Sunrise Instruments) enable the measurement of needle shaft forces and operator inputs, respectively. Note, that shaft forces are only measured for comparison. During the insertion, the movement is restricted to the axial needle direction. To include haptic feedback into the position control we employ the control loop depicted in Fig.~\ref{fig:control}. The inner control loop is governed by the position controller of the fast research interface~\cite{schreiber2010fast}. In the outer control loop an implicit force trajectory-tracking controller similar to~\cite{Roy.2002} is implemented. The robot position 
\begin{equation}
    x_d(t) = k_i \int_0^t e_f(t) \, dt + k_p e_f(t)
\end{equation}
is choosen such that the error $e_F = {F_{\text{H}}} - \alpha {F_{\text{T}}}$ between the handle force ${F_{\text{H}}}$ and the amplified measured tip force $F_{\text{T}}$ is minimized. The magnitude $e_F$ is limited between $0$ and $F_{\text{H}}$ to prevent involuntary movement.
The gain $\alpha$ can be chosen by the operator. The control loop and robot communication is implemented with the Robot Operating System (ROS). Force measurements and haptic control run at a frequency of \SI{200}{\Hz}.

\subsection{\textit{Smart} Needle}
We build custom needle probes with an integrated optical force sensor which we refer to as \textit{smart} needle. Our \textit{smart} needle is integrated into an introducer needle of a clinical biopsy system and allows sensing of forces at the needle tip.
The \textit{smart} needle components are depicted in Fig.~\ref{fig:needle}, left. A symmetrical needle tip is guided by the needle sheath with an outer diameter of~\SI{2.05}{\milli\meter}. An inner tube centers the optical fibre within the needle sheath. The fiber is cleaved for common-path imaging and placed within~\SI{0.5}{\milli\meter} of the proximal end of the needle tip's guide. The compression layer between needle tip and sheath causes the OCT signal to change under load. OCT data is recorded with a spectral domain OCT system (Telesto Telesto I, Thorlabs GmbH, GER). The system records one-dimensional depth scans (A-scans) with a maximum imaging depth of approximately \SI{2.6}{\milli\meter} in air resolved over 512 pixels.

For real-time data processing, we consider convolutional gated recurrent units (cGRU) with a subsequent 1D CNN designed for spatio-temporal input data (Fig.~\ref{fig:needle}, right).Similar to Gessert et al.~\cite{gessert2019spatio}, we replace the dot products in the GRU cells with 1D convolutions such that 
\begin{align*}
    z_t &= \sigma ( W_{hz} * h_{t-1} + W_{xz} * x_{t} + b_z ), \\
    r_t &= \sigma ( W_{hr} * h_{t-1} + W_{xr} * x_{t} + b_r ), \\
    \hat{h}_t &= tanh( W_{h} * (r_t \odot h_{t-1}) + W_{x} * x_{t} + b) \text{ and} \\
    h_t &= (1 - z_t) \odot h_{t-1} + z_t \odot \hat{h}_t
\end{align*}
defines the update gate $z_t$, the reset gate $r_t$, the candidate activation vector $\hat{h_t}$ and the hidden state $h_t$, respectively. By updating the trainable filers $W$, spatial information is processed for each A-scan and temporal information is extracted in the recurrent unit. The tip force is estimated based on the resulting feature vector $h_n$ in the regression head (Fig.~\ref{fig:needle}). We compare our cGRU-CNN model with a basic 2D residual neural network (ResNet)~\cite{resnet2016deep}.

To calibrate our \textit{smart} sensor prior to insertion, we manually apply cyclic axial loads on a rigid surface. We record \SI{6e4}{} synchronized OCT A-scans and force labels for tuning our model, with forces between \SIrange{0}{5}{\newton}. 
We train our models on input sequences of 50 A-scans over 50 epochs with a learning rate of \num{5e-4} and a batch size (B) of 128, using the mean squared error (MSE) as our loss function. We use Adam optimization with default parameters~\cite{kingma2014adam}. We test the model employment over \SI{1e4}{} A-scans implementing a cyclic buffer to maintain the input dimensions for the ResNet model. We compare the two architectures with regard to accuracy and check for real time applicability. We report mean absolute errors (MAE) and Pearson correlation coefficient (pCC). To evaluate real time application, we report timings for forward and backward pass and the inference time for individual samples processed on the GPU (RTX 3070, NVIDIA Corporation, USA). The model with the lowest MAE during calibration is used for phantom and in-situ insertions.

\subsection{Phantom Experiments}
\begin{figure}[tb]
  \centering
  \includegraphics[ width=0.99\columnwidth]{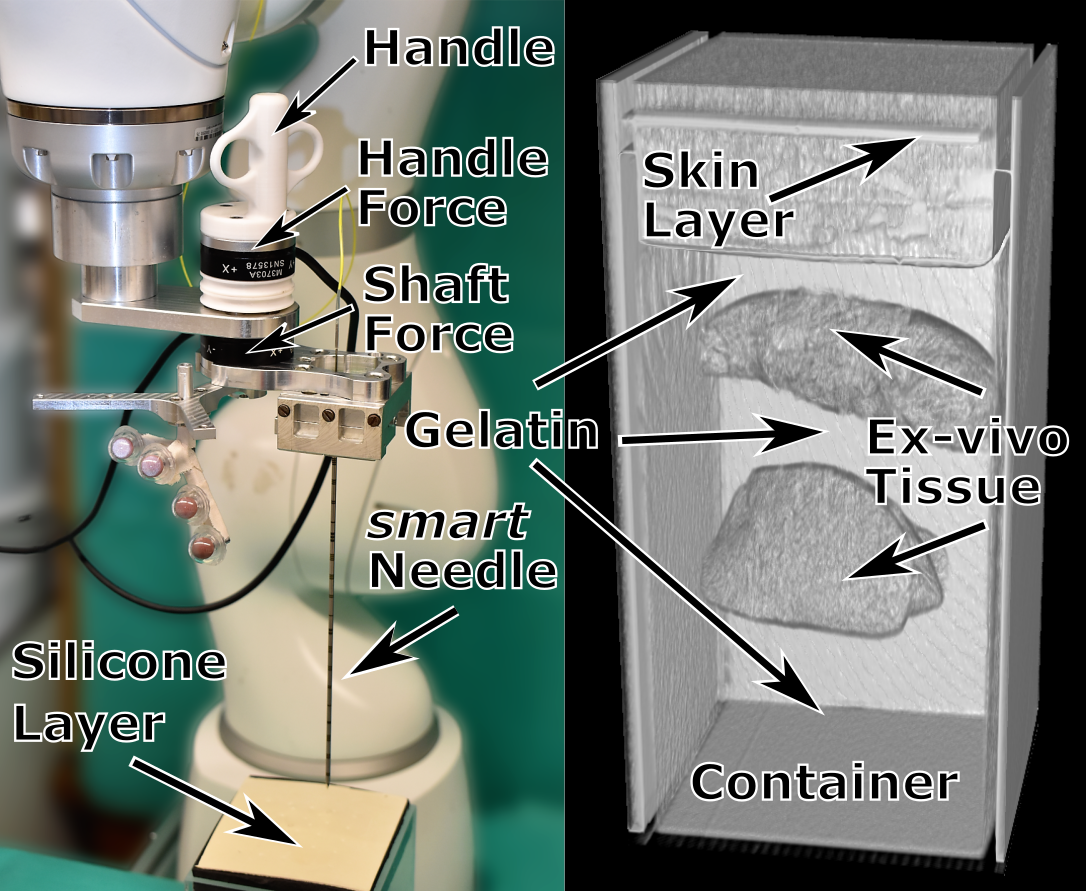}
  \caption{\textbf{Phantom Experiments:} The \textit{smart} needle and the haptic system with regard to tissue interface detection are evaluated in a phantom study. The system (left) allows for both fully robotic and collaborative insertions. (Right) CT scan of a phantom containing ex-vivo tissue embedded in gelatin to prevent bulk displacement. A skin layer with silicone and polyethylene foam emulates friction forces from the skin.}
  \label{fig:phantom}
\end{figure}

We perform a phantom study to evaluate if the needle tip forces are beneficial in the detection of tissue interfaces. For this purpose, we embed ex-vivo chicken muscle tissue into gelatin gels that fixate the tissue and prevent bulk displacement (Fig.~\ref{fig:phantom}). A skin layer at the top of the phantom consists of polyethylene foam and silicone rubber to simulate the superposition of friction forces in percutaneous insertions. We manufacture in total four phantoms and acquire CT scan to determine the location of interfaces based on the measured Hounsfield units. 
Firstly, we perform 25 fully robotic insertions with a constant velocity of \SI{5}{\milli\meter\per\second} to evaluate needle forces independent of operator inputs. We compare absolute forces at the needle tip with the location of tissue interfaces marked in the pre-insertion CT scan. 
To underline the importance of local tip force measurements, we additionally analyze friction forces by subtracting the tip force from the total axial force. We report changes in friction force per material{ }\textemdash{ }skin layer, gelatin and ex-vivo tissue{ }\textemdash{ }by calculating the slope of a linear regression through the friction forces of each segment separated by tissue interfaces.

Secondly, we conduct a user study in which five participants are tasked to sense interfaces during collaborative needle insertions. We provide kinesthetic feedback on tip forces and the participants are tasked to enable a trigger if a tissue interface is perceived. We report the distance between the estimated needle tip position triggered by the user and the position of the tissue interface in the CT reference frame. We evaluate the number of correctly detected and missed tissue interfaces. We distinguish between entry and exit events for each of the two tissue layers, resulting in up to four marked locations per insertion.  
Participants perform three test insertions to choose the gain $\alpha$ with which the tip forces are scaled. 

\subsection{In Situ Pancreas Biopsy}
\begin{figure}[tb]
  \centering
  \includegraphics[ width=0.90\columnwidth]{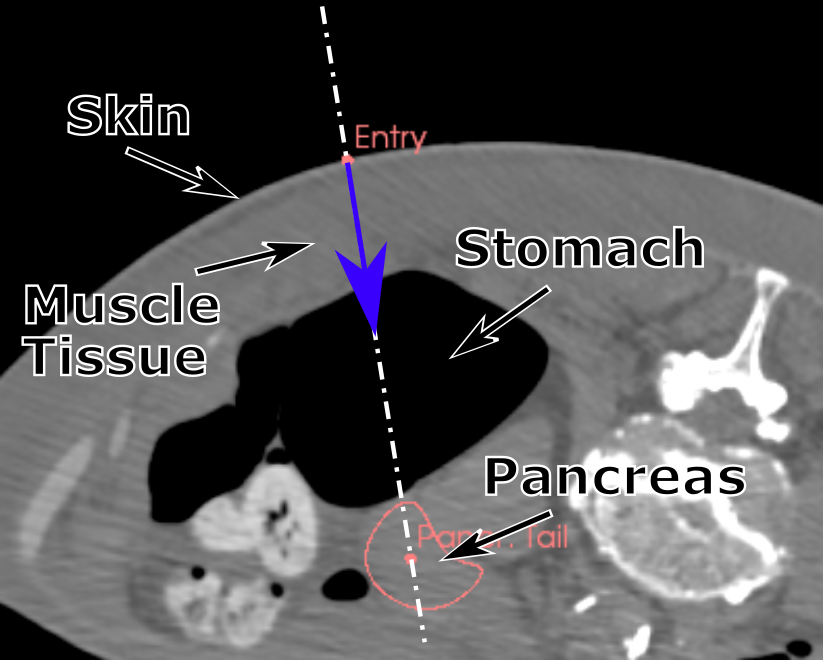}
  \caption{\textbf{Pancreas Biopsy:} Biopsy of the tail of the pancreas is performed in a collaborative approach. Image guidance is used to globally align the needle along the planned trajectory (white dashed line) based on the selection of the surgeon. Local needle placement is performed by the surgeon controlling axial motion along the trajectory (blue arrow). }
  \label{fig:biopsy_plan}
\end{figure}

We demonstrate the collaborative system in a forensic setup by performing post-mortem pancreatic biopsy in two different cases. The target area is marked by an expert in the pre-insertion CT and a path through the center of the pancreatic tail is planned for robotic trajectory guidance (Fig.~\ref{fig:biopsy_plan}). The needle is pre-aligned by the robot and the insertion is switched to collaborative control. An incision is made into the skin in order to reduce forces upon dermal entry and the pathologist controls the placement of the needle. The pathologist performs collaborative needle insertion and stops the motion once he felt the tissue transition of the target structure. With the needle sheath held in place, the \textit{smart} needle is retracted and a sample is taken with the biopsy gun at the chosen position. A post-insertion CT is acquired to visualize final needle placement.

\section{Results}
In the following, we report the accuracy of our needle tip force calibration. We then present our ex-vivo tissue study and finally demonstrate the application of our system for post-mortem pancreas biopsy. 

\subsection{\textit{Smart} Needle Calibration}
\begin{figure}[t]
  \centering
  \includegraphics[trim={0.0cm, 0cm, 0cm, 0cm}, clip, width=0.9\columnwidth]{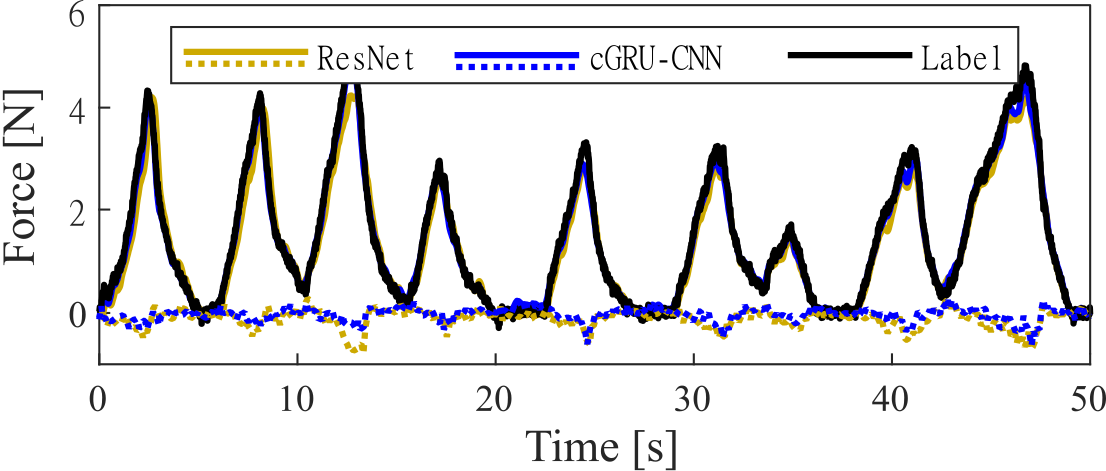}
  \caption{\textbf{\textit{Smart} Needle Calibration:} Example plot of the needle tip force calibration with estimations for both model architectures and the ground truth measurements (black). The absolute errors for both models are displayed as dotted lines.}
  \label{fig:calibration}
\end{figure}

Needle tip force estimations for both models are reported in Tab.~\ref{tab:models}. While the total time per sample for the forward and backward pass is longer for the recurrent model, samples can be processed independently during use and both models can be integrated into the control loop running at \SI{200}{\Hz}. The spatio temporal cGRU-CNN model outperforms the ResNet architecture with a MAE of \SI{0.11}{\newton} and the more accurate model is consequently chosen during phantom and post-mortem insertions. An example of the calibrated tip force estimation during loads exclusively applied at the tip can be seen in Fig.~\ref{fig:calibration}.

\begin{table}[t]
    \centering
    \caption{Error metrics and timings for the network architectures. The inference time (IT) and the total time (TT) for forward and backward pass are given for individual examples.}   
    \begin{tabular}{l c c c c}
    \toprule
       Model  & MAE [N] & pCC & IT [ms] & TT [ms] \\ \hline
    ResNet  & 0.15 & 0.99 & 0.30 & 15.13\\
    cGRU-CNN  & 0.11 & 0.99  & 1.03 & 52.96 \\\bottomrule
    \end{tabular}
    \label{tab:models}
\end{table}

\begin{figure}[t]
  \centering
  \includegraphics[trim={0.0cm, 0cm, 0cm, 0.0cm}, clip, width=0.95\columnwidth]{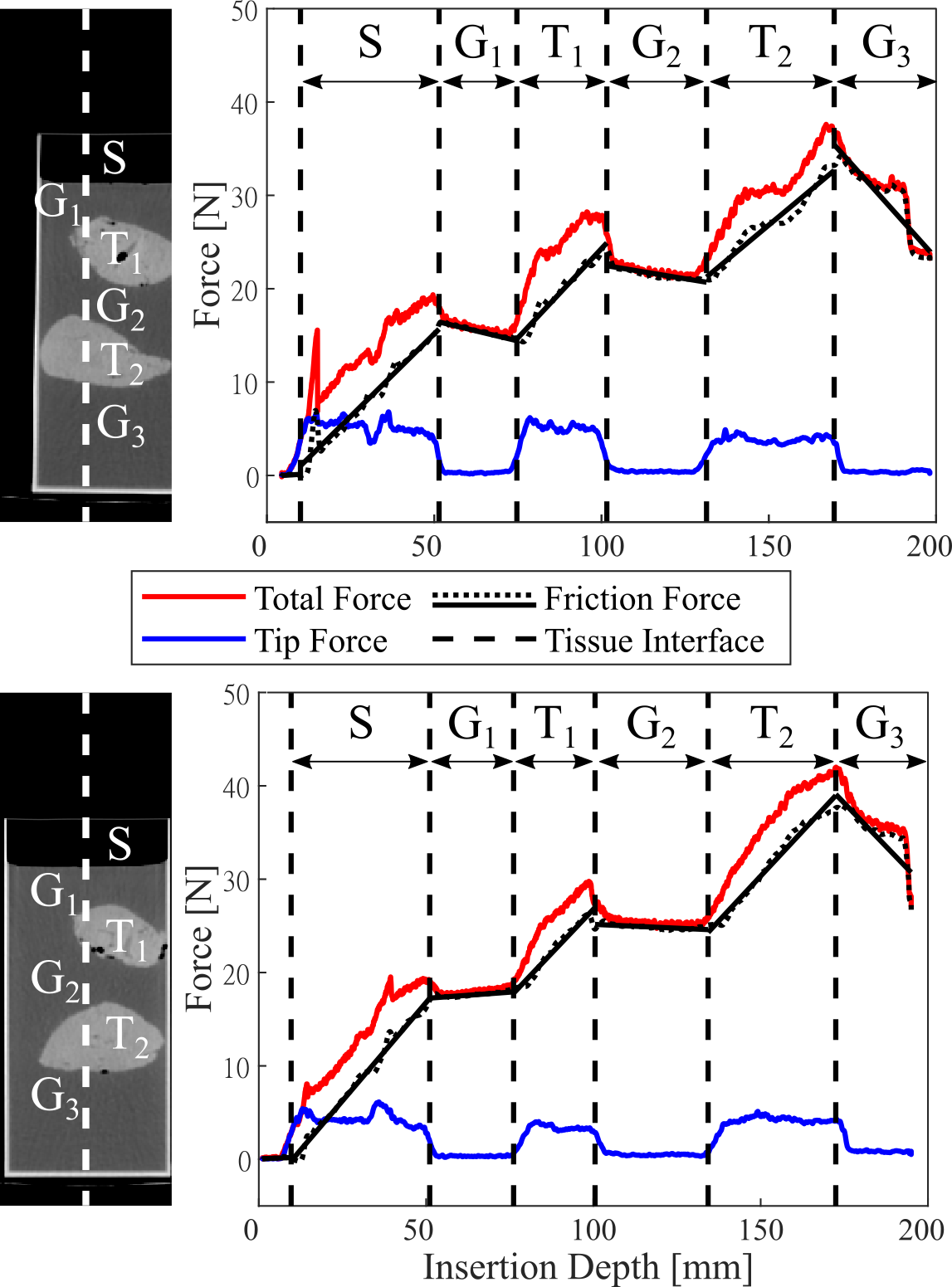}
  \caption{\textbf{Phantom Needle Insertions:} Two examples (top and bottom) for insertions with constant velocity through ex-vivo tissue with pre-insertion imaging and trajectory (dleft) and corresponding force plots (right). Location of tissue interfaces (dashed), tip forces (blue), total forces measured at the shaft (red), friction forces (dotted) and linear regression of friction forces (black) are plotted over the insertion depth. Insertions are segmented into the topmost skin segment (S), gelatin segments (G) and tissue segments (T). }
  \label{fig:phantom_forces}
\end{figure}

\begin{figure}[t]
\centering
	\includegraphics[width = 0.95\columnwidth]{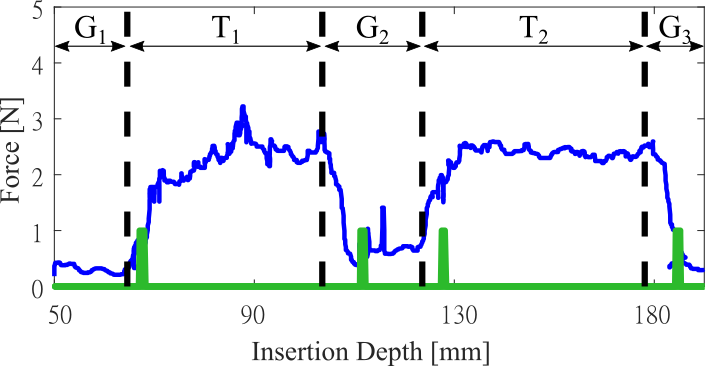}
   \caption{\textbf{Example Needle Insertion with Haptic Feedback:} Depicted are the needle tip forces (blue) and trigger (green) enabled manually by the user when a tissue transition is sensed. The position of the gelatin layers (G) and tissue layers (T) are indicated on top.}
  \label{fig:phantom_forces_c}
\end{figure}

\subsection{Phantom Experiments}
Two examples of the robotic insertions into ex-vivo tissue with constant velocity can be seen in Fig.~\ref{fig:phantom_forces}.
During 25 automatic insertions, an increase in tip force $F_{\text{T}}>\SI{1}{\newton}$ can be noted \SI{2.08(108)}{\milli\meter} below the entry into ex-vivo tissue marked in the pre-insertion CT scan. Similarly, the return to tip forces $F_{\text{T}}<\SI{1}{\newton}$ corresponding to the re-entry into homogeneous gelatin occurs \SI{5.13(188)}{\milli\meter} below the locations marked prior to the insertion. We further report the friction force per unit length in Tab.~\ref{tab:friction_slopes} separated by skin layer (S), gelatin (G) and ex-vivo tissue (T). Large variations over different segments of the same material can be seen. Especially for segments where the needle tip is cutting through homogeneous gelatin, friction forces vary between \SI[per-mode=repeated-symbol]{0.05}{\newton\per\milli\meter} and \SI[per-mode=repeated-symbol]{-0.56}{\newton\per\milli\meter}. Examples of this can be seen in Fig.~\ref{fig:phantom_forces}, indicated by decreasing (e.g. $\text{G}_1$, top) and increasing slopes (e.g. $\text{G}_1$, bottom) in gelatin for different insertions. 

\begin{table}[tb]
    \centering
    \caption{Mean, standard deviation and range for friction per unit length [\SI[per-mode=repeated-symbol]{}{\newton\per\milli\meter}] for each material.}
    \begin{tabular}{cccc}

    \toprule
      Material & Skin Layer & Tissue & Gelatin \\\midrule
      Mean & \SI{0.38(4)}{}  & \SI{0.36(8)}{} & \SI{-0.15(19)}{}\\
      Min & \SI{0.25}{}  & \SI{0.13}{} & \SI{-0.56}{}\\
      Max & \SI{0.45}{}  & \SI{0.54}{} & \SI{0.05}{}\\
     \bottomrule
    \end{tabular}
    \label{tab:friction_slopes}
\end{table}

The results of the user study can be seen in Tab.~\ref{tab:insertionsPhantom} where participants are tasked to feel for the topology of the phantoms.
The mean distance between tissue interfaces estimated from CT and user inputs ranges from \SI{5.77}{\milli\meter} to \SI{11.74}{\milli\meter} for all participants. For the interfaces from gelatin to tissue layers the mean distance for all performed experiments is smaller (\SI{5.45(331)}{\milli\meter}) compared to interfaces between tissue and gelatin (\SI{9.85(489)}{\milli\meter}). The distance to interfaces from gelatin to tissue are \SI{4.85(244)}{\milli\meter} and \SI{6.05(402)}{\milli\meter} for the two tissue layers, respectively. For the interfaces from tissue to gelatin we report increased distances of \SI{9.14(423)}{\milli\meter} and \SI{10.88(575)}{\milli\meter} for the two tissue layers, respectively. In total, users do not recognize 9 tissue interfaces resulting in a total detection rate of \SI{91}{\percent}. Of the missed cases, \SI{80}{\percent} correspond to the interface between tissue and gelatin.

\setlength{\tabcolsep}{0.35em}
\begin{table}[tb]
    \caption{Detection rate (DR) and distances [$\SI{}{\milli\meter}$] between pre-insertion position and positions detected by the user for the five participants.}   
\begin{tabular}{lccccc}
\toprule
DR & $\text{G}_1$ to $\text{T}_1$ & $\text{T}_1$ to $\text{G}_2$ &  $\text{G}_2$ to $\text{T}_2$ &  $\text{T}_2$ to $\text{G}_3$ & Mean \\ \midrule
\SI{95}{\percent}    & \SI{2.86(150)}{} & \SI{10.23(449)}{}  & \SI{5.95(210)}{}  & \SI{10.94(390)}{}  & \SI{7.31(445)}{}    \\
\SI{80}{\percent}    & \SI{7.21(147)}{} & \SI{14.30(257)}{}  & \SI{11.86(714)}{}  & \SI{14.90(580)}{}  &  \SI{11.74(497)}{}    \\
\SI{95}{\percent}    & \SI{4.20(336)}{} & \SI{6.46(409)}{}  & \SI{3.35(174)}{}  & \SI{12.80(866)}{}  &  \SI{6.38(570)}{}    \\
\SI{90}{\percent}    & \SI{4.85(167)}{} & \SI{7.89(288)}{}  & \SI{4.14(287)}{}  & \SI{6.47(383)}{}  &  \SI{5.77(295)}{}    \\
\SI{95}{\percent}    & \SI{5.12(220)}{} & \SI{6.81(178)}{}  &  \SI{7.27(263)}{} &  \SI{8.63(347)}{} &  \SI{6.77(253)}{}    \\ \midrule
Mean & \SI{4.85(244)}{} & \SI{9.14(423)}{}  &  \SI{6.05(402)}{} &  \SI{10.88(575)}{} &     \\ \bottomrule 

\end{tabular}
    \label{tab:insertionsPhantom}
\end{table}

\subsection{In Situ Pancreas Biopsy}
\begin{figure}[t]
  \centering
  \includegraphics[trim={0cm, 0cm, 0cm, 0cm}, clip, width=\columnwidth]{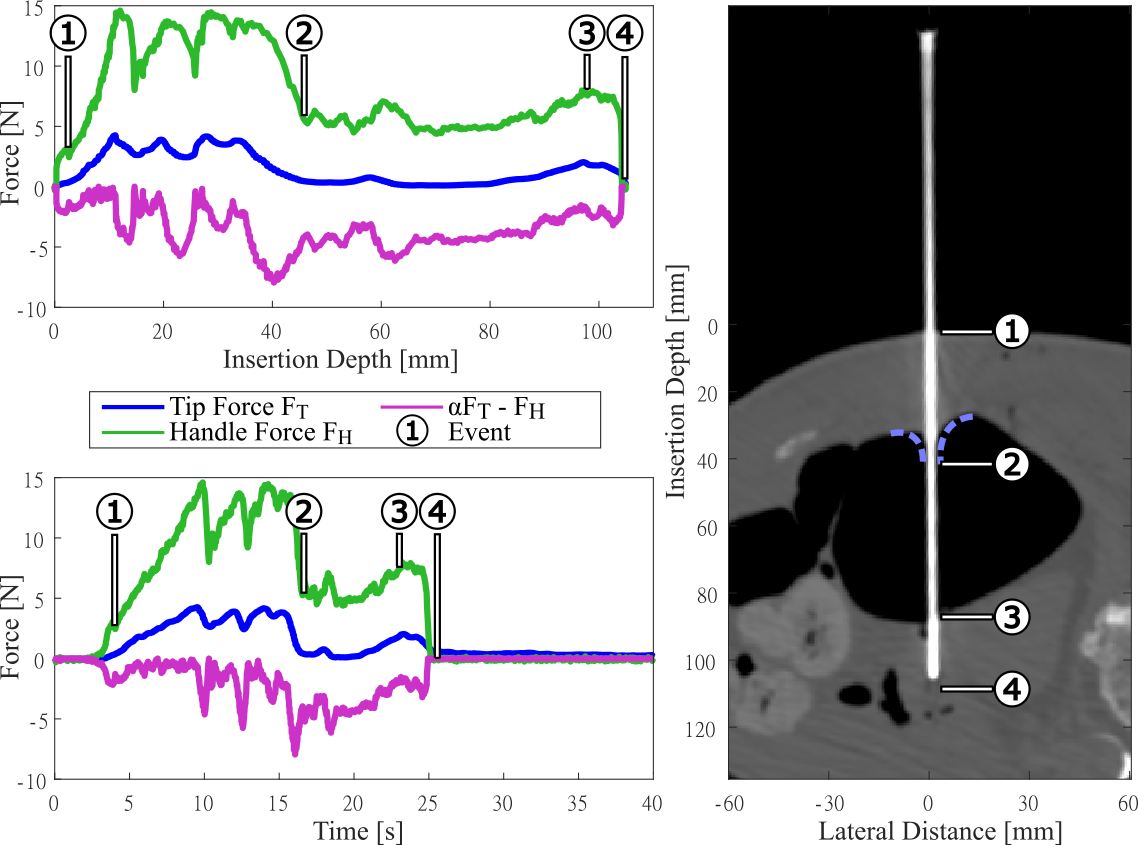}
  \caption{\textbf{Haptic Guided Pancreas Biopsy:} Needle tip forces (blue), axial handle force (green) and the difference $\alpha {F_{\text{T}}} - {F_{\text{H}}}$ (magenta) displayed over the insertion depth (top, left) and insertion time (bottom, left). Corresponding CT scan after the insertion (right) with marked events denoting percutaneous entry (1), entry (2) and exit (3) of the stomach and final needle position (4). Inward deformation of the stomach wall (dashed blue line) causes the rupture (2) to occur at a greater depth than can be seen on the pre- or post-insertion CT scan.}
  \label{fig:haptic_insertion} 
\end{figure}

\begin{figure}[t]
    \centering
    \subfloat[]{
        \includegraphics[trim={1.5cm, 0cm, 1.5cm, 0cm}, clip, height=30mm]{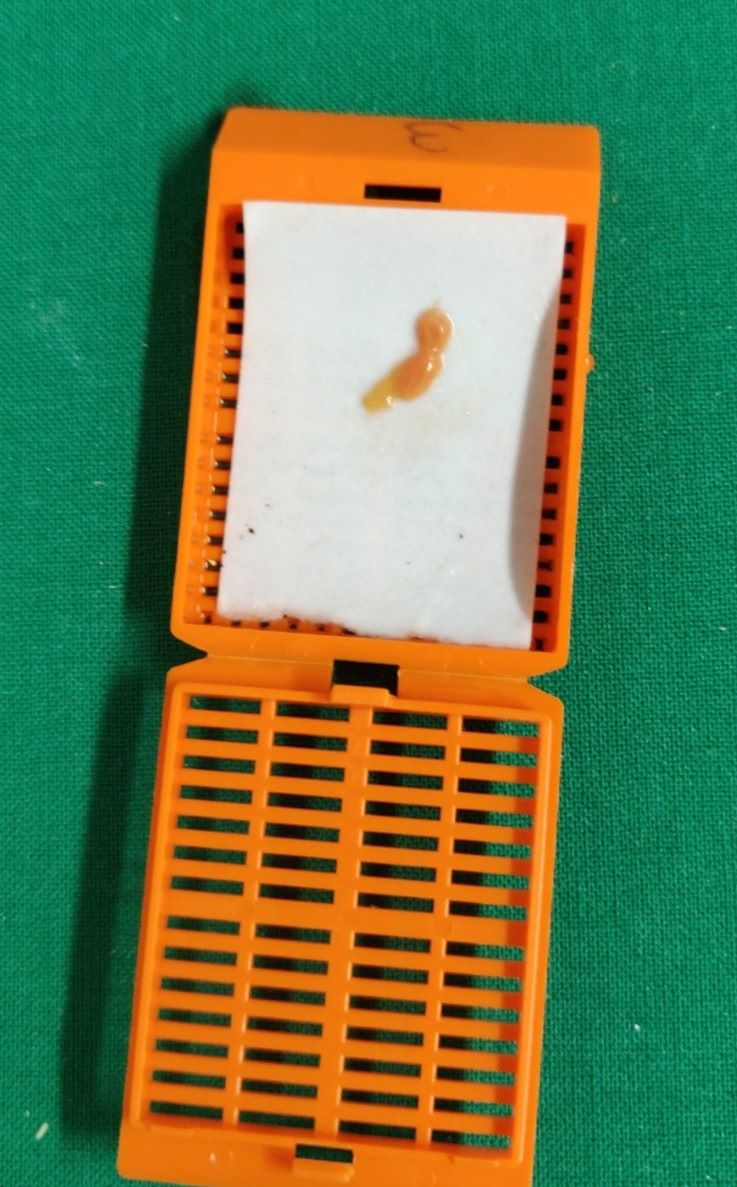}
        \label{fig:pancreas_sample}
    }%
    \subfloat[]{
        \includegraphics[trim={2.6cm, 0cm, 2.6cm, 0cm}, clip, height=30mm]{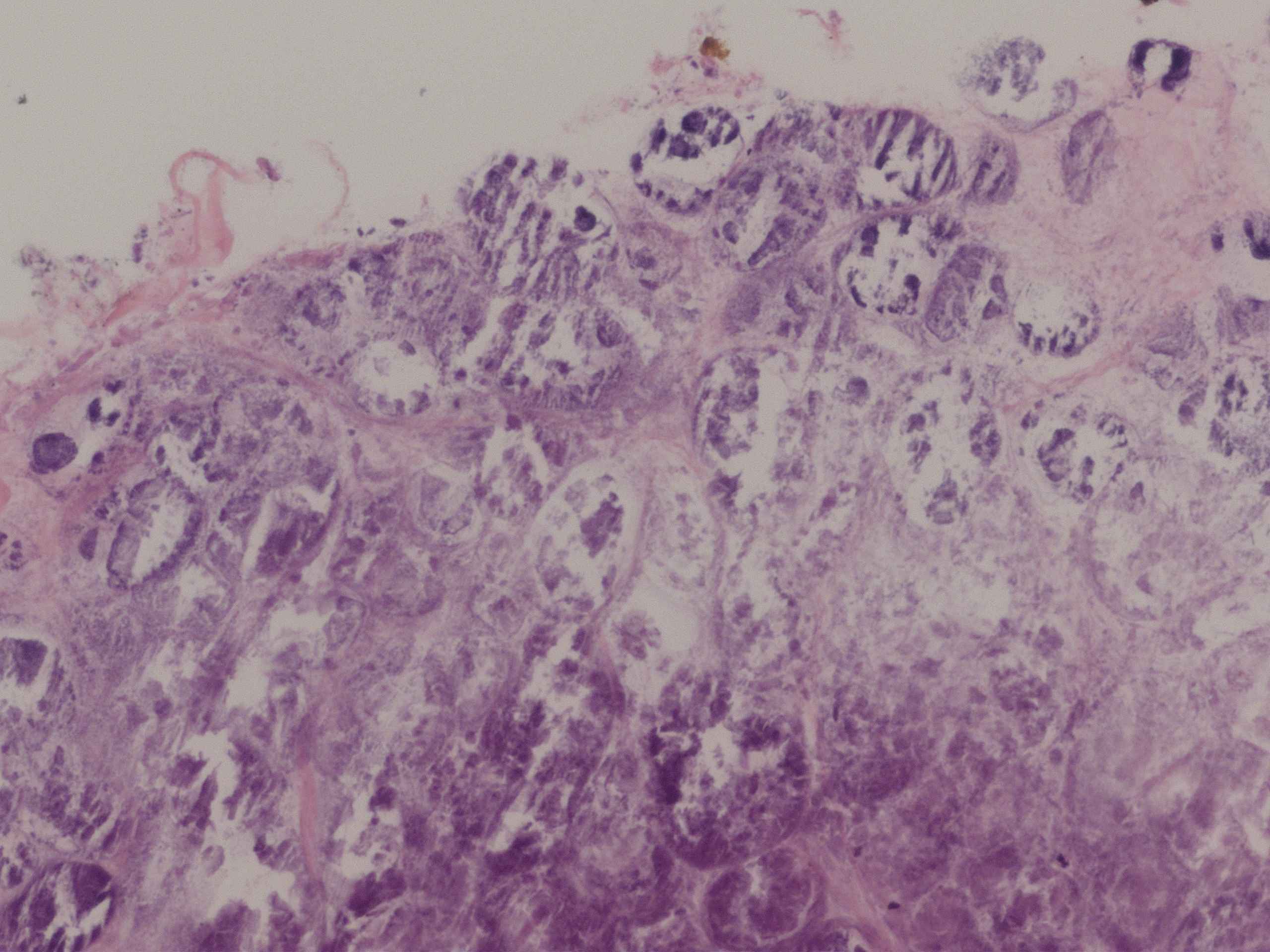}
        \label{fig:pancreas_histo}
    }%
    \subfloat[]{
        \includegraphics[trim={2cm, 0cm, 2cm, 0cm}, clip, height=30mm]{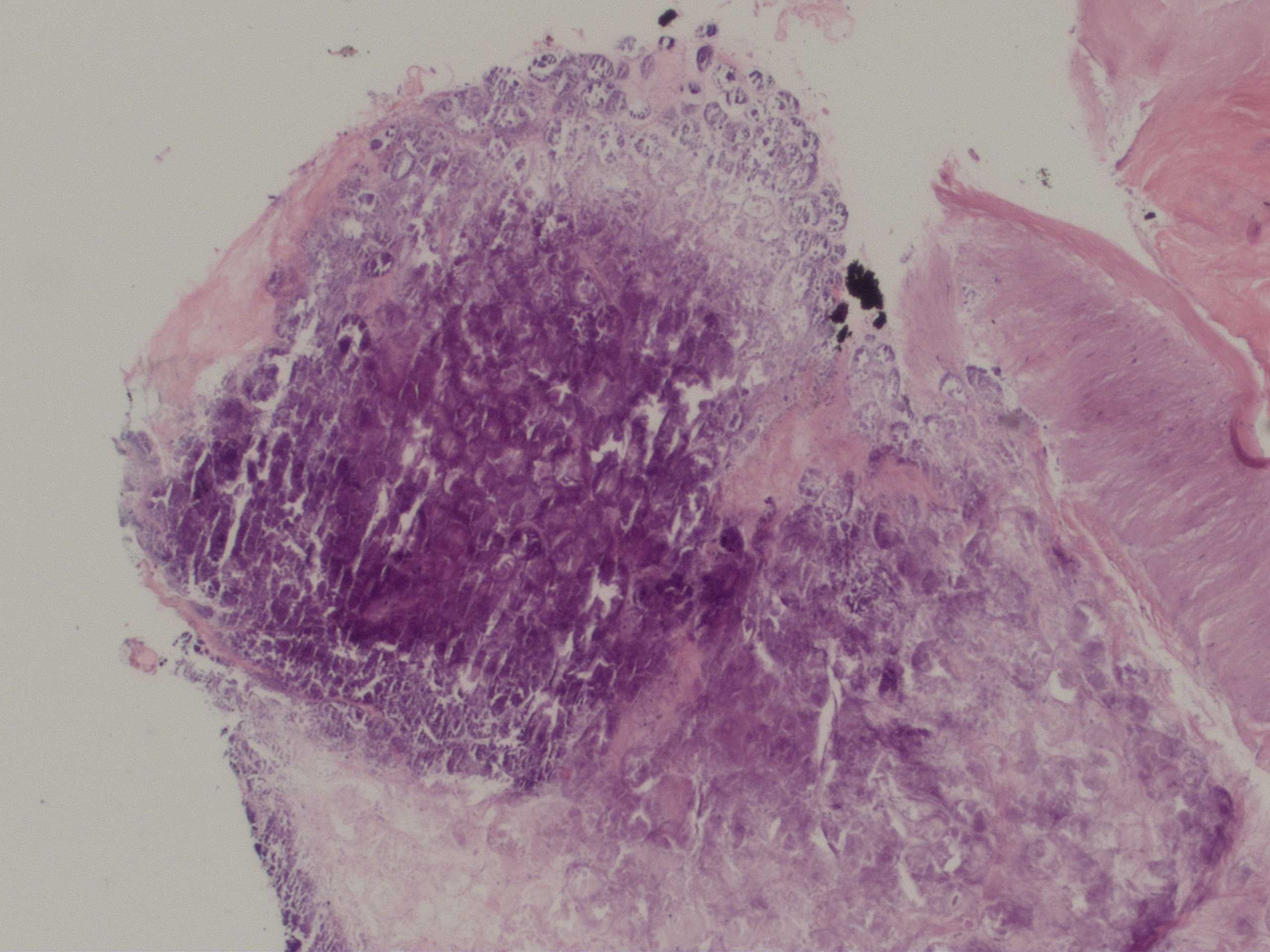}

        \label{fig:pancreas_histo_b}
    }
    \caption{\textbf{Pancreatic Tissue Biopsy:} Extracted sample (a) corresponding to the insertion shown in Fig.~\ref{fig:haptic_insertion}) and histopathological imaging (b) confirming the success of the pancreatic biopsies in both cases (although tissue is already affected by putrefaction).}
    \label{fig:sample_and_histo}
\end{figure}

A visualization of a collaboratively performed pancreatic biopsies can be seen in Fig.~\ref{fig:haptic_insertion}. With the needle aligned along the planned trajectory, the pathologist advances the needle through the initial resistance of skin and muscle tissue and passes into the stomach (2). As the needle exits the partially gas-filled stomach the pathologist detects the force peak and subsequent rupture (3) and is able to place the needle in the tail of the pancreas. The samples taken are subjected to a histopathological examination, which confirms the successful insertion (Fig.~\ref{fig:sample_and_histo}). 

\section{Discussion and Conclusion}
In this work, we present a concept for a collaborative robotic biopsy system that combines trajectory guidance with real-time feedback of needle tip forces. Haptic feedback is provided by the robot on site, enabling the physician to locate tissue interfaces during insertion.

We perform ex-vivo tissue experiments for validation and systematic analysis of estimated forces acting on the needle tip and the corresponding kinesthetic feedback.
Insertions with a constant velocity show that friction forces per unit length vary within the same material even in a phantom setup with ex-vivo tissue embedded in gelatin. In Fig.~\ref{fig:phantom_forces} the friction forces show either falling (top) or rising slopes (bottom) within the homogeneous gelatin and the inconsistent behavior is most likely due to the mechanics of relaxation in tissue~\cite{mahvash2009mechanics, okamura2004force}. Consequently, the friction forces are not only strongly dependent on the insertion velocity, tissue properties and needle geometry, but also on the combination of already punctured tissue.
Our findings underline the benefit of directly obtaining absolute tip force values as demonstrated in~\cite{ elayaperumal2014detection, de2012coaxial, han2018mr} and highlight the limitations of friction model based approaches~\cite{aggravi2021haptic}. We have observed that the sharp tip of our \textit{smart} needle results in nearly constant tip forces between tissue interfaces cutting through each layer and limiting tissue compression. 

The user study shows that the provided kinesthetic feedback was interpreted more reliably by the participants for increasing forces, corresponding to needle punctures into tissue (T). Tissue to gelatin interfaces were detected with less accuracy. This can be partially explained by the design of the needle as approximately \SI{5}{\milli\meter} of the needle tip protrudes beyond the needle sheath (Fig.~\ref{fig:needle}) and a fraction of the load remains until the tip is fully extended beyond the tissue interface. In addition, gelatin prevents bulk displacement, but limited tissue deformation can still occur especially during the transition from one layer to the next. Adjustments to the needle design e.g. a shorter tip could further enhance detection as resistance decreases. The current system is limited to kinesthetic feedback that the users must first learn to interpret. Additional sensations, e.g. vibrotactile feedback could help improve detection rates and reduce user dependence~\cite{aggravi2021haptic}. Variability between naive users shows that the haptic feedback is not equally intuitive to every participant with detection rates ranging from \SI{80}{\percent} to \SI{95}{\percent}. In comparison, the detection of membrane puncture events in a phantom setup were reported with a \SI{75}{\percent} success rate in~\cite{elayaperumal2014detection} and \SI{98.9}{\percent} in~\cite{han2018mr}. In~\cite{de2012coaxial}, success rates ranged from \SI{50.0}{\percent} to \SI{83.3}{\percent}.
However, comparisons to previous works are challenging as they are highly dependent on tissue and experimental setup, e.g. the perception of membrane puncture with uni-axial motion stages~\cite{han2018mr}. 

Lastly, we demonstrate collaborative robotic biopsy in a real-word scenario. The pancreas represents a challenging target for biopsies within the retroperitoneal space as ultrasound image guidance is hampered for imaging deep tissues~\cite{malek2005percutaneous}. Manual insertions with robotic trajectory guidance can assist in needle placement~\cite{Guiu.2021, Levy.2021}. Here we provide haptic feedback on tissue interfaces to the physician, potentially increasing needle placement accuracy. The pathologist was able to perceive the interface between stomach and pancreas, as indicated in~Fig.\ref{fig:haptic_insertion}, and successfully position the needle relative to this target structure. The biopsy of the pancreas demonstrates the feasibility of the approach for anatomically difficult located tissue. But this approach would also be suitable for other soft tissue biopsy targets, e.g. lung or prostate. The in-situ application demonstrates the needle tip sensor under realistic load but sensitivity under strong lateral forces needs to be further evaluated. Our initial results with a small sample size and a single operator are promising, but further evaluations regarding applicability and clinical workflow integration need to be explored in the future.

In conclusion, our results suggest that haptic feedback is a valuable alternative to fully automated robotic needle placement. Our results demonstrate that it is possible to sense tissue interfaces with a collaborative robot for versatile needle insertions in a large operating space. With our system the physician is at all times in control of the needle insertion which is crucial in a clinical environment. Further studies will show how collaborative robotic biopsy compares to exclusive trajectory guidance and manual placement, and how much training is required for the adaptation processes.






\section*{ACKNOWLEDGMENT}
The authors state no conflict of interest.

\bibliographystyle{IEEEtran}
\bibliography{refs}

\end{document}

%% file: figs/control_scheme.tex
\tikzstyle{block} = [draw, rectangle, 
    minimum height=2em, minimum width=3em]
\tikzstyle{sum} = [draw, circle] 
\tikzstyle{input} = [coordinate]
\tikzstyle{midpoint} = [coordinate]

\begin{tikzpicture}[auto, node distance=2 cm, every node/.style={font=\small}] 
    \node [input, name=op]{};
    \node[sum, right =0.8cm of op] (sum1){};
    \node[block, right=0.4cm of sum1](pid){PI};
    \node[sum, right =0.7cm of pid] (sum2){};
    \node[block, right =0.4cm of sum2, text width = 1.1cm](fri){FRI Control};
    \node[sum, right =0.7cm of fri](sum3){};
    \node[block, right =0.4cm of sum3](robot){Robot};
    \node[coordinate, right = 0.3cm of robot] (q_s) {};
    \node[coordinate, below = 0.3cm of q_s] (q_m) {};
    \node[block, below=0.6cm of robot](env){Env.};
    
    \node[block, left = 0.6cm of env](sensor){Needle Tip Force Sensor};
    \node[block, left = 0.6cm of sensor](gain){$\alpha$};

    \node[block, above = 0.5cm of sum2](kin){Kinematics};
    \node[midpoint, right =0.6cm of op] (Fh){};

    \draw[draw, ->](op) -- node[near start, font=\bfseries] {$\mathbf{F_{\text{H}}}$} node[pos=0.92] {$+$} (sum1);
    \draw[->] (sum1) -- node {$e_{F}$}(pid);
    \draw[->] (pid) -- node[name=qd]{$x_d$} node[pos=0.92] {$+$} (sum2);
    \draw[->] (sum2) -- node {$e_{x}$}(fri);
    \draw[->] (fri) -- node[near start] {$\tau_{D}$} node[pos=0.92] {$+$} (sum3);
    \draw[->] (sum3) -- node {} (robot);

    \draw[->] (Fh) |- node {} (kin);
    \draw[->] (kin) -| node[near end] {$\tau_{F}$} node[pos=0.92] {$+$}  (sum3);

    \draw[->] (robot) -- node[name=q]{$x$} (env);
    \draw[->] (q) -| node [pos=0.92] {$-$} (sum2);
    \draw[->] (env) -- (sensor); 
    \draw[->] (sensor) -- node [above, font=\bfseries] {$\mathbf{F_{\text{T}}}$} (gain);
    \draw[->] (gain) -| node[pos=0.92] {$-$}  (sum1);

\end{tikzpicture}